\documentclass[letterpaper]{article} 
\usepackage{aaai2027}  
\usepackage[hyphens]{url}  
\usepackage{graphicx} 
\usepackage{natbib}  
\usepackage{caption} 
\usepackage{algorithm}
\usepackage{algorithmic}

\usepackage{newfloat}
\usepackage{listings}
\DeclareCaptionStyle{ruled}{labelfont=normalfont,labelsep=colon,strut=off} 
\floatstyle{ruled}
\newfloat{listing}{tb}{lst}{}
\floatname{listing}{Listing}

\usepackage{booktabs}
\usepackage{multirow}
\usepackage{makecell}
\usepackage{amssymb}

\title{Latent Variable-Mediated Cross-Learning for Few-Shot Acoustic Impedance Imaging}
\author {
    Junheng Peng\textsuperscript{\rm 1},
    Yong Li\textsuperscript{\rm 1} \corresponding,
    Mingwei Wang\textsuperscript{\rm 2}, 
    Yi Bao\textsuperscript{\rm 3}
}
\affiliations {
    \textsuperscript{\rm 1}The Key Laboratory of Earth Exploration \& Information Techniques of Ministry Education, Chengdu University of Technology, Chengdu, 610059, China\\
    \textsuperscript{\rm 2}School of Resources and Safety Engineering, Chongqing University, Chongqing, 400044, China\\
    \textsuperscript{\rm 3}Institute of Space Earth Science, Nanjing University, Suzhou, 215163, China\\
    JunhengPeng@stu.cdut.edu.cn, Liyong07@cdut.edu.cn
}

\begin{document}

\maketitle

\begin{abstract}

Acoustic impedance imaging is a fundamental yet severely ill-posed problem in subsurface analysis: the seismic wavelet is unknown, observations are band-limited, and labeled well-log samples are extremely scarce (typically $<$1$\%$ of all traces). Existing semi-supervised deep learning methods mitigate few-shot problem by incorporating forward modeling, yet they either rely on inaccurate prior wavelet assumptions or introduce auxiliary networks, leading to unstable optimization and degraded performance. We propose RD-SCL, a novel framework that integrates regularized deconvolution with semi-supervised cross-learning. At its core lies a differentiable, closed-form first-order Tikhonov deconvolution operator that dynamically estimates the latent wavelet in the frequency domain during training, providing stable physics-guided feedback without explicit auxiliary networks and fixed wavelet priors. Building on this operator, we design a symmetric cross-learning that enforces consistency between predictions on labeled and unlabeled data, thereby effectively exploiting abundant unlabeled traces. Extensive experiments on the SEAM and Marmousi 2 benchmarks demonstrate that RD-SCL consistently outperforms state-of-the-art supervised and semi-supervised methods, achieving substantial gains with lower computational cost. With only 56.5k learnable parameters and competitive runtime, RD-SCL offers a practical, physically consistent, and efficient solution for acoustic impedance imaging.

\end{abstract}

\begin{links}
    \link{Example code}{https://github.com/lexiaoheng/RD-SCL}
\end{links}

\section{Introduction}

Acoustic impedance imaging, also known as acoustic impedance inversion (AII), is a fundamental yet challenging task in geophysical exploration. Acoustic impedance can characterize variations in subsurface lithology, porosity, and structure \cite{1, 2}, providing indispensable references for understanding subsurface and resource development \cite{3, 4}. 

\begin{figure*}[t]
\centering
\includegraphics[width=\textwidth]{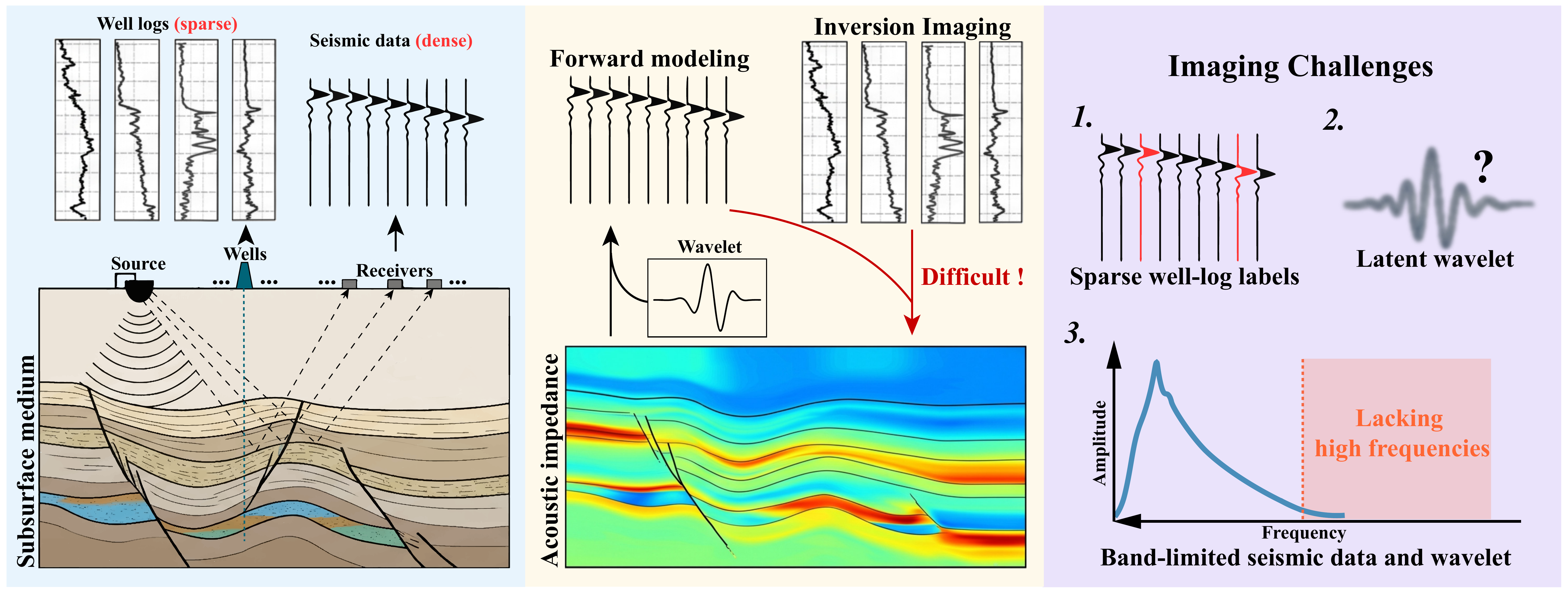} 
\caption{The acquisition of dataset, background, and challenges of acoustic impedance imaging.}
\label{fig: background}
\end{figure*}

Performing subsurface AII using well logging is extremely expensive, while a more cost-effective approach combines surface seismic data with sparse well-logs for inversion, as shown in Fig.\ref{fig: background}. However, this inverse task is fundamentally ill-posed and nonlinear: (i) due to the extreme sparsity of well-log data (less than 1$\%$ of the total data), AII is a few-shot task; (ii) the seismic wavelet is unknown and hard to estimate without strong priors \cite{6}; (iii) seismic signals is band-limited, concentrating energy at low frequencies \cite{7}, so direct deconvolution cannot recover high-frequency components. 

With the advancement of artificial intelligence, deep learning has been introduced into the field of seismic processing \cite{8}. For AII, deep learning offers a powerful data-driven alternative \cite{9}. However, purely data-driven methods often behave as "black boxes" and struggle to preserve the physical information of seismic wave propagation with limited labels \cite{10, 11}, leading to either geologically implausible artifacts or severe performance degradation \cite{12}. Moreover, effectively leveraging the vast amount of unlabeled data remains a critical challenge for AII, as labeled data are typically extremely sparse and expensive in field exploration \cite{13, 14}. 

Previous deep learning AII methods typically impose predefined priors during training or use multi-network for consistency constraints. In contrast, we propose an latent variable-mediated mechanism that dynamically estimates latent wavelet from the predictions and transfers them between labeled and unlabeled traces to establish bidirectional consistency. First, inspired by classical regularization \cite{15} for direct inversion, we construct a differentiable frequency-domain deconvolution operator that dynamically estimates the latent variable from the predicted impedance during training. Second, using this closed-form operator, we build a self-correcting physics-driven feedback loop, in which the inversion results for abundant unlabeled data are constrained. Finally, we develop a cross-learning paradigm that enforces mutual consistency between labeled and unlabeled data, enabling stable optimization of the deep learning network. The main contributions are as follows: 
\begin{itemize}
    \item We design a regularized deconvolution operator based on Tikhonov regularization, which enables efficient, stable, and closed-form dynamic latent variable estimation in the frequency domain and can be integrated into the network training iterations.
    \item We propose a cross-learning framework that integrates the above operator, establishing mutual constraints between labeled and unlabeled data and effectively alleviating the overfitting and label-sparsity issues prevalent in AII.
    \item Extensive experiments on benchmarks demonstrate that our method significantly improves inversion accuracy and physical consistency, achieving substantial gains over state-of-the-art methods.
\end{itemize}

\section{Related Works}
\subsection{Acoustic Impedance Imaging}
Early optimization schemes primarily relied on iterative least-squares methods \cite{16}, which require a reliable initial low-frequency model to constrain the solution space. To overcome the local minima that plague least-squares methods, heuristic algorithms were subsequently introduced, including simulated annealing \cite{17, 30}, ant colony optimization \cite{18}, genetic algorithms \cite{19}, and particle swarm optimization \cite{20}. Although these conventional frameworks are theoretically rigorous and have laid the foundation for modern seismic characterization, they suffer from critical practical limitations: they incur intensive computational overhead, are highly sensitive to parameter tuning, and depend heavily on prior assumptions about the seismic wavelet \cite{21}. Due to these limitations, such methods require repeated experiments and case-specific parameter tuning for different datasets, limiting their general applicability and efficiency.

\subsection{Supervised Methods}
Deep learning approaches to AII can be broadly divided into supervised and semi-supervised paradigms. Early supervised frameworks utilized temporal convolutional networks (TCNs) \cite{9, 10}, recurrent neural networks (RNNs) \cite{50}, and multi-scale residual structures \cite{11} to train inverse networks directly on traces adjacent to the wells. To address the limitations of trace-by-trace networks, encoder-decoder architectures have been introduced to map seismic traces uniformly into a high-dimensional feature space \cite{22}. However, purely supervised methods often lack interpretability and yield limited inversion accuracy and tend to overfit to local patterns when labeled data are scarce, motivating the development of semi-supervised alternatives.   

\subsection{Semi-Supervised Methods}
To address the limitations of supervised networks on unlabeled data, semi-supervised and physics-informed deep learning frameworks have emerged as a dominant research direction \cite{12, 23}. These methods typically integrate a forward modeling operator into the training loop, enforcing that the impedance predictions on unlabeled traces, when passed through the forward network, reconstruct the observed seismic data \cite{21, 24, 31, 34}. Physics-driven approaches leverage the physical convolutional model of seismic wave propagation \cite{25, 29}, but their success remains highly contingent on accurate extraction of the seismic wavelet. In contrast, network-driven semi-supervised methods involve the sequential training of two deep networks \cite{26, 27, 28, 35, 36}; for unlabeled data, however, the training objectives of the two networks may diverge, leading to degraded inversion accuracy.

\begin{figure*}[t]
\centering
\includegraphics[width=\textwidth]{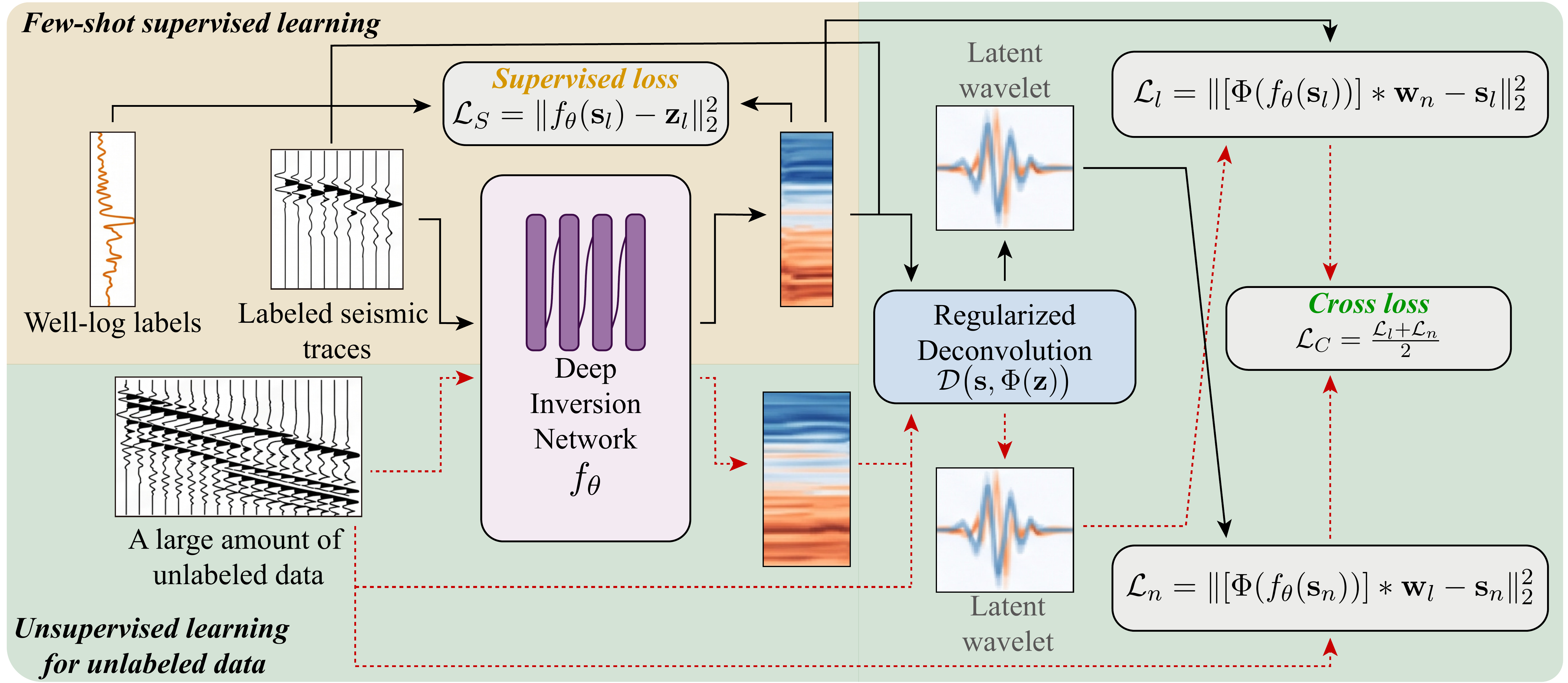} 
\caption{The proposed regularized deconvolution and acoustic impedance imaging workflow.}
\label{fig: workflow}
\end{figure*}

\section{Methodology}

In this section, we propose regularized deconvolution and cross-learning framework for AII. Unlike previous semi-supervised paradigms that rely on a secondary network to approximate the seismic forward process, we treat the wavelet as a latent variable and propose an explicit dynamic estimation operator; building upon this, we construct a semi-supervised cross-learning strategy to achieve unsupervised alignment for unlabeled data, which maintains strict physical consistency. 

\subsection{Forward Modeling}

Seismic data $\mathbf{S} \in \mathbb{R}^{T \times N}$ are vibration signals acquired along the surface at certain spatial intervals, as shown in Fig.\ref{fig: background}; each seismic trace $\mathbf{s}_i \in \mathbb{R}^{T}$ is a time series, which can be converted to the depth domain through time-depth model. Besides, the modeling of seismic trace \cite{5} as: 
\begin{equation}
\label{eq:r to s}
\mathbf{s}_i = \mathbf{w} \ast \mathbf{r}_i + \mathbf{\epsilon}_i= \int_{0}^{T} \mathbf{w}(\tau) \, \mathbf{r}_i(t - \tau) \, d\tau + \mathbf{\epsilon}_i,
\end{equation}
where $\mathbf{w} \in \mathbb{R}^{T_w}$ is the seismic wavelet, $\mathbf{\epsilon}_i$ is random noise, and $\mathbf{r}_i \in \mathbb{R}^{T}$ is the reflection coefficient computed from the acoustic impedance:
\begin{equation}
\label{eq:z to r}
\mathbf{r}_{i,t} = \Phi(\mathbf{z_{i, t}})= \frac{\mathbf{z}_{i,t+dt} - \mathbf{z}_{i,t}}{\mathbf{z}_{i,t+dt} + \mathbf{z}_{i,t}}, \quad t = 1, \dots, T-1.
\end{equation}
Thus, AII aims to recover the full impedance $\mathbf{Z} \in \mathbb{R}^{T}$ from seismic traces $\mathbf{S}$ with only sparse well-log labels. In addition, the definition of acoustic impedance and the derivation of the forward process based on the wave equation can be found in \textit{Supplementary Materials}). 

\subsection{Regularized Deconvolution}

Rather than relying on auxiliary networks or fixed priors to approximate the latent wavelet $\mathbf{w}$, we treat it as a latent variable and derive a physical estimator via regularized deconvolution in the frequency domain. This operator forms the cornerstone of our cross-learning framework, enabling bidirectional consistency constraints between labeled and unlabeled traces.

Concretely, given an input seismic trace $\mathbf{s}_i$ and the network's predicted impedance $\mathbf{z}_i$, we first compute the corresponding reflection coefficient $\mathbf{r}_i$ via Eq.~(\ref{eq:z to r}). We then transform both quantities into the frequency domain:
\begin{equation}
    \tilde{\mathbf{r}}_i(\omega) = \int_{0}^{T} \mathbf{r}_i(t) e^{-i\omega t} dt,
\end{equation}
\begin{equation}
    \tilde{\mathbf{s}}_i(\omega) = \int_{0}^{T} \mathbf{s}_i(t) e^{-i\omega t} dt.
\end{equation}

Applying the convolution theorem to Eq.~(\ref{eq:r to s}), the time-domain convolution reduces to element-wise multiplication:
\begin{equation}
\label{eq:freq_conv}
\tilde{\mathbf{s}}_i(\omega) = \tilde{\mathbf{r}}_i(\omega) \cdot \tilde{\mathbf{w}}_i(\omega) + \tilde{\mathbf{\epsilon}}_i(\omega),
\end{equation}
where $\tilde{\mathbf{w}}_i(\omega)$ and $\tilde{\mathbf{\epsilon}}_i(\omega)$ denote the frequency-domain representations of the latent wavelet and random noise, respectively. Here, $\tilde{\mathbf{\epsilon}}_i(\omega)$ naturally absorbs both observational noise and the prediction error incurred by the deep learning network during training.

Solving for $\tilde{\mathbf{w}}_i$ is is a textbook ill-posed inverse problem: at frequencies where $|\tilde{\mathbf{r}}_i(\omega)|$ approaches zero, even modest noise in $\tilde{\mathbf{s}}_i(\omega)$ is catastrophically amplified, producing wildly oscillatory variable estimates that destabilize the entire training pipeline. To robustly estimate the latent variable while preserving differentiability, we employ first-order Tikhonov regularization in the frequency domain:
\begin{equation}
\label{eq:tikhonov}
\min_{\tilde{\mathbf{w}}_i} \mathcal{J}(\tilde{\mathbf{w}}_i) = \big\| \tilde{\mathbf{r}}_i(\omega) \cdot \tilde{\mathbf{w}}_i(\omega) - \tilde{\mathbf{s}}_i(\omega) \big\|_2^2 + \lambda \big\| \mathbf{L} \tilde{\mathbf{w}}_i \big\|_2^2,
\end{equation}
where $\mathbf{L}$ is the first-order differential operator. In the time domain $\mathbf{L} = \frac{d}{dt}$, and in the frequency domain $\mathbf{L} = i\omega$, making the penalty term $\lambda \omega^2 |\tilde{\mathbf{w}}_i|^2$. This choice of regularizer is physically well-motivated: seismic wavelets are predominantly low-frequency, so penalizing high-frequency energy effectively suppresses noise amplification while preserving the wavelet's spectral support.

For each independent frequency $\omega$, $\mathcal{J}$ is a real-valued function of the complex variable $\tilde{\mathbf{w}}_i$. Setting the Wirtinger derivative to zero, $\frac{\partial \mathcal{J}}{\partial \tilde{\mathbf{w}}_i^*} = 0$, yields the normal equation:
\begin{equation}
-\tilde{\mathbf{r}}_i^* \cdot (\tilde{\mathbf{s}}_i - \tilde{\mathbf{r}}_i \cdot \tilde{\mathbf{w}}_i) + \lambda \, \omega^2 \, \tilde{\mathbf{w}}_i = 0.
\end{equation}

Solving for $\tilde{\mathbf{w}}_i$ yields the closed-form, per-frequency estimate:
\begin{equation}
\label{eq:wiener}
\tilde{\mathbf{w}}_i(\omega) = \frac{\tilde{\mathbf{s}}_i(\omega) \cdot \tilde{\mathbf{r}}_i^*(\omega)}{|\tilde{\mathbf{r}}_i(\omega)|^{2} + \lambda \, \omega^2}.
\end{equation}
The time-domain latent variable is then recovered via the inverse Fourier transform:
\begin{equation}
\mathbf{w}_i(t) = \frac{1}{2\pi} \int_{-\infty}^{+\infty} \frac{\tilde{\mathbf{s}}_i(\omega) \cdot \tilde{\mathbf{r}}_i^*(\omega)}{|\tilde{\mathbf{r}}_i(\omega)|^{2} + \lambda \, \omega^2} \, e^{i\omega t} d\omega.
\end{equation}

We summarize the overall procedure as the deconvolution operator $\mathcal{D}$:
\begin{equation}
\label{eq:decon}
\mathbf{w} = \frac{1}{\mathcal{B}} \sum_{i=1}^\mathcal{B} \mathbf{w}_i = \mathcal{D}\big(\mathbf{s}, \Phi(\mathbf{z})\big),
\end{equation}
where $\mathcal{B}$ denotes the batch size. Crucially, batch-level averaging yields a more stable estimate than trace-wise estimation, as individual prediction errors are partially uncorrelated across traces and are attenuated through aggregation.

Based on ablation experiments and prior work~\cite{32}, we set $\lambda = 0.01$ across all experiments. We note that estimating $\mathbf{r}$ from $\mathbf{w}$ and $\mathbf{s}$ is substantially harder due to the band-limited nature of both quantities, which further motivates our design choice of estimating the wavelet as the latent bridge variable rather than solving the full deconvolution end-to-end. 
\begin{figure*}[t]
\centering
\includegraphics[width=\textwidth]{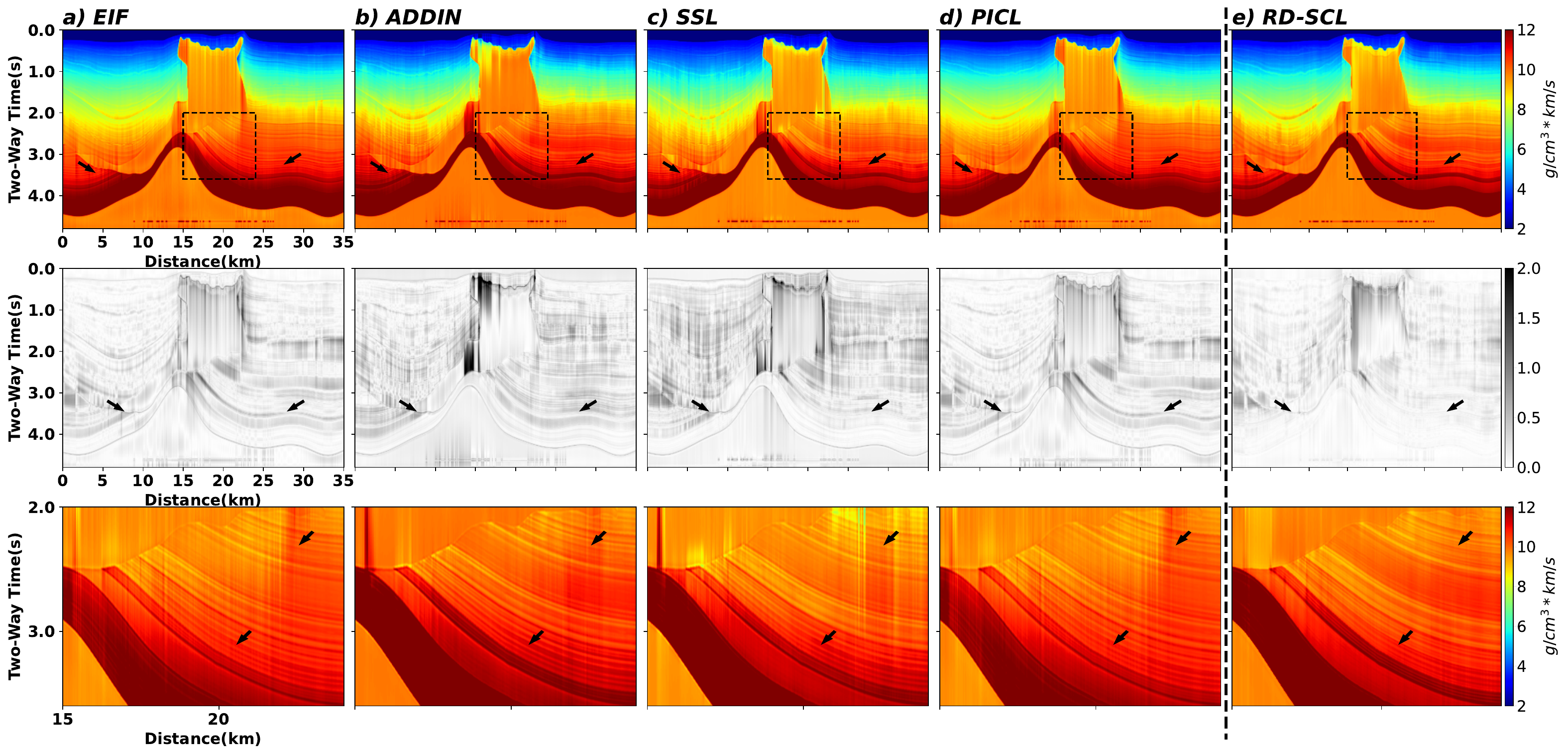} 
\caption{The results of EIF, ADDIN, SSL, PICL, and RD-SCL on SEAM dataset. The second row shows the absolute residuals between each result and the true acoustic impedance; the third row is a magnification of the region outlined by the black box.}
\label{fig: SEAM}
\end{figure*}
\begin{figure*}[t]
\centering
\includegraphics[width=\textwidth]{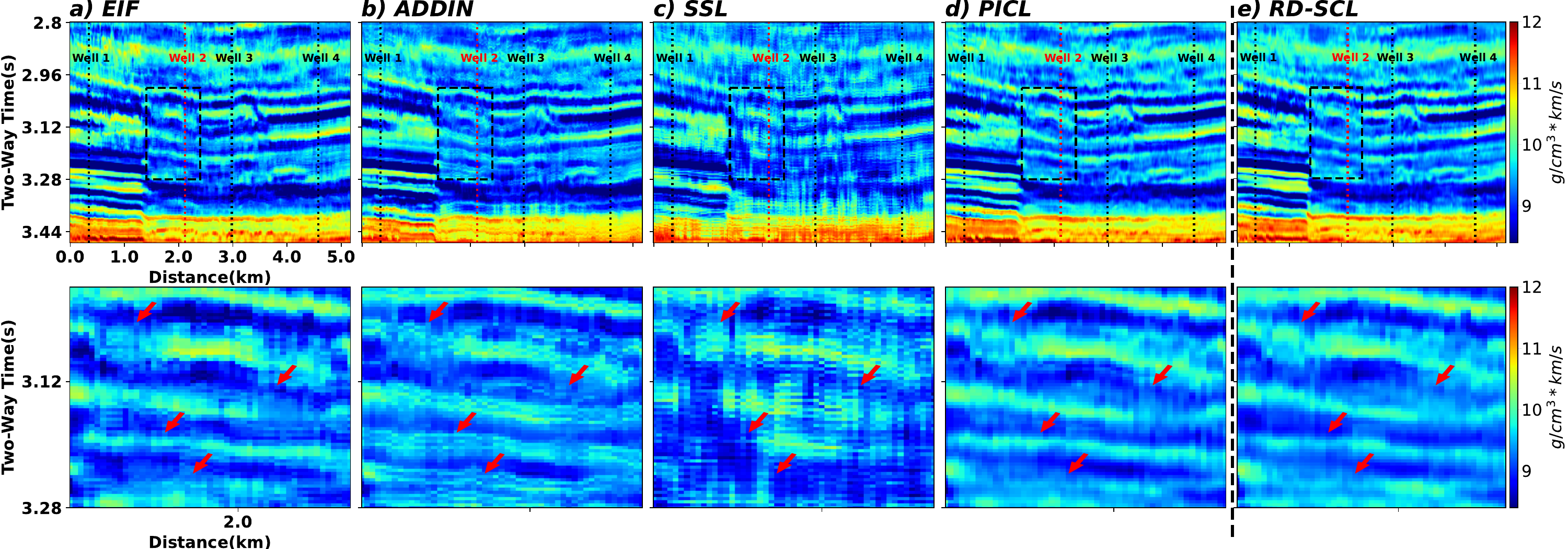} 
\caption{The results of EIF, ADDIN, SSL, PICL, and RD-SCL on field dataset, while the second row is a magnification of the region outlined by the black box. Well 2 is the blind well used for testing and not involved in training. }
\label{fig: field}
\end{figure*}
\begin{figure}[t]
\centering
\includegraphics[width=\columnwidth]{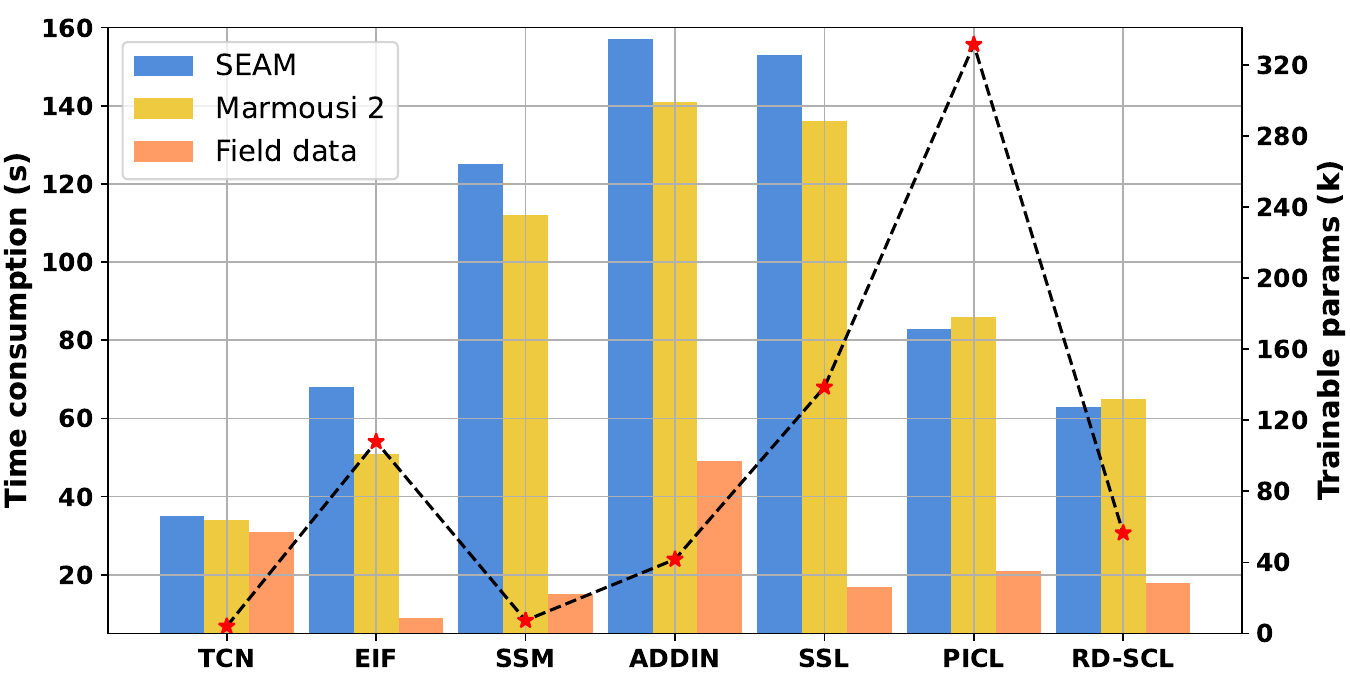} 
\caption{The computation time required by each method on three datasets, along with the number of trainable parameters.}
\label{fig: efficiency}
\end{figure}
\begin{figure}[t]
\centering
\includegraphics[width=\columnwidth]{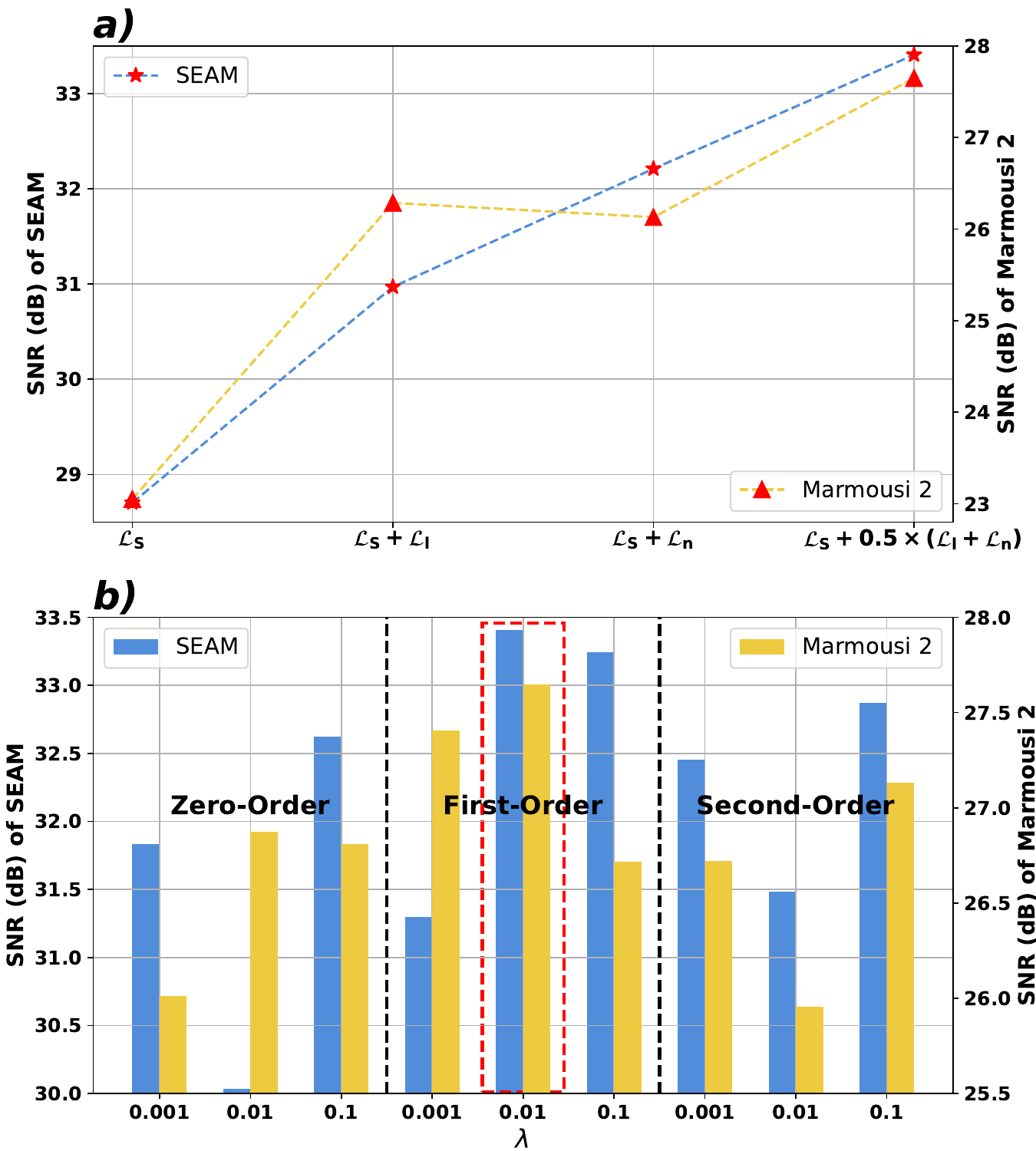} 
\caption{(a) shows the contributions of each component proposed in this work; (b) shows the SNRs of the processing results under zero-order, first-order, and second-order Tikhonov regularization with different $\lambda$.}
\label{fig: ablation}
\end{figure}

\subsection{Semi-Supervised Cross-Learning}

To address the few-shot learning challenge, we introduce a cross-learning strategy combined with regularized deconvolution. Unlike existing semi-supervised AII methods that either rely on auxiliary networks \cite{26,27,28} or depend on pre-estimated wavelet priors~\cite{21,24,25}, the proposed framework employs only a single deep neural network $f_\theta$, parameterized by $\theta$, and dynamically estimates the latent wavelet through the differentiable operator $\mathcal{D}$ during training. For any input seismic trace $\mathbf{s}_i$, the network directly predicts the continuous acoustic impedance:
\begin{equation}
\hat{\mathbf{z}}_i = f_\theta(\mathbf{s}_i).
\end{equation}
The training target is composed of two cooperative components: supervised loss and cross loss.

\subsubsection{Supervised Loss}
On the labeled dataset, the network parameters $\theta$ are optimized to minimize the discrepancy between the predicted impedance and the ground-truth well logs (Fig.~\ref{fig:  workflow}):
\begin{equation}
\mathcal{L}_{S} = \| f_\theta(\mathbf{s}_l) - \mathbf{z}_l \|_2^2.
\end{equation}
where $\mathbf{s}_l$ and $\mathbf{z}_l$ denote the labeled data and labels, respectively, which typically account for only a very small fraction of the entire dataset. 

\subsubsection{Cross Loss}
For the large amount of unlabeled data $\mathbf{s}_n$, we adopt a cross-learning strategy that treats the wavelet as a latent bridge variable: rather than directly enforcing prediction consistency between labeled and unlabeled branches, we transfer the dynamically estimated wavelets across them, thereby establishing physically-grounded bidirectional consistency constraints without introducing auxiliary networks. Concretely, we first perform regularized deconvolution on the output of the labeled and unlabeled data:
\begin{equation}
\mathbf{w}_l = \mathcal{D}(\mathbf{s}_l, \Phi(\mathbf{\hat{z}}_l)),
\end{equation}
\begin{equation}
\mathbf{w}_n = \mathcal{D}(\mathbf{s}_n, \Phi(\mathbf{\hat{z}}_n)).
\end{equation}
Due to the supervised loss, $\mathcal{D}(\mathbf{s}_l, \Phi(\mathbf{\hat{z}}_l))$ and $\mathcal{D}(\mathbf{s}_l, \Phi(\mathbf{z}_l))$ are approximately identical. For unlabeled data, we adopt the forward loss as the unsupervised objective:
\begin{equation}
\mathcal{L}_{n} = \| [\Phi(f_\theta(\mathbf{s}_n))] * \mathbf{w}_l - \mathbf{s}_n \|_2^2.
\end{equation}
Symmetrically, we define:
\begin{equation}
\mathcal{L}_{l} = \| [\Phi(f_\theta(\mathbf{s}_l))] * \mathbf{w}_n - \mathbf{s}_l \|_2^2.
\end{equation}
The cross-learning loss is then $\mathcal{L}_{C} = \frac{\mathcal{L}_{l}+\mathcal{L}_{n}}{2}$, as illustrated in Fig.~\ref{fig: workflow}. Importantly, this formulation delivers unsupervised learning for unlabeled data, effectively mitigating the overfitting risk inherent in few-shot tasks. 

In summary, the overall training objective combines supervised and cross-learning losses:
\begin{equation}
\mathcal{L}(\theta) = \mathcal{L}_{S} + \mathcal{L}_{C},
\end{equation}
which we minimize via gradient descent for semi-supervised training. 

\subsection{Network Structure}
In AII, the network architecture is not the primary determinant of accuracy; moreover, given the scarcity of labels, a more complex network may exacerbate overfitting. The network adopted in this work has a deliberately simple architecture: it consists of four temporal convolutional network (TCN) blocks, three bidirectional gated recurrent unit (Bi-GRU) layers, and one linear layer, and detailed architecture is presented in \textit{Supplementary Materials}). The hidden dimensions of each layer are $[16, 16, 16, 32, 32, 32, 32]$. Each TCN block comprises two dilated convolutional layers with kernel size 3 and dilation factor 2, interleaved with tanh activations, and is designed to capture long-range sequential dependencies. The total number of learnable parameters is approximately 56.5k, striking a balance between lightweight design and modeling capacity.

\section{Experiments and Discussions}
\subsection{Setup}
\subsubsection{Datasets}
We use two synthetic datasets for validation: the SEAM elastic earth model \cite{33} and Marmousi 2 \cite{34}. Constructed from field survey lines of 35 km and 13.6 km in length, respectively, both have served as standard benchmarks in AII research over the past decade. The number of labeled samples accounts for less than 1$\%$ of the total data, which is far lower than that in comparable works. In addition, to evaluate the application potential of proposed method, we use an oil and gas exploration and production dataset from the Sichuan Basin, China. Compared with the synthetic benchmarks, the field data exhibit more complex geological structures and higher noise levels. The detailed dataset parameters are presented in \textit{Supplementary Materials}).

\subsubsection{Compared Methods and Metrics} 
We refer to our proposed method as Regularized Deconvolution Semi-Supervised Cross-Learning (RD-SCL), and compare it against six open-source baselines. For supervised learning, we include TCN \cite{10} and EIF \cite{22}; for semi-supervised learning, we include SSM \cite{35}, ADDIN \cite{28}, SSL \cite{36}, and PICL \cite{12}. All methods are deployed in the same environment with parameter settings kept consistent with their respective original publications and public codes. We evaluate all methods using five metrics: signal-to-noise ratio (SNR), structural similarity (SSIM), coefficient of determination (R$^2$), mean absolute error (MAE), and mean squared error (MSE). Before computing MAE and MSE, we apply Z-score normalization to the predictions. 

\subsubsection{Environment}
During training, we use a fixed learning rate of 0.003, a batch size of 6, and train for 1,000 epochs. Each dataset is trained and evaluated independently without transfer learning, and all hyperparameters are kept identical across datasets. We adopt the AdamW optimizer with a weight decay of 0.01. In terms of the runtime environment, we use an Intel Core i5-14400F CPU with 16 GB of RAM and an RTX 4060 GPU with 8GB RAM for deep learning acceleration. All comparison methods and RD-SCL are implemented using PyTorch 2.4.1 and run under unified random seeds (five sets) to ensure experimental reproducibility.

\subsection{Comparison on Synthetic Datasets}
\begin{table*}[!t]
\centering
\scriptsize
\setlength{\tabcolsep}{1.2mm}
\renewcommand\arraystretch{1.2}
\begin{tabular}{cc|cccccc|c}
\toprule[1.2pt]
\textbf{Metric} & \textbf{Dataset}  & \textbf{TCN} & \textbf{EIF} & \textbf{SSM} & \textbf{ADDIN} & \textbf{SSL} & \textbf{PICL} & \textbf{RD-SCL}\\
\cline{1-9}
\multirow{2}{*}{\textbf{SNR $\uparrow $}} & SEAM & $24.18\pm3.26$ & $30.32\pm1.54$ & $24.75\pm2.17$ & $26.89\pm2.43$ & $26.44\pm2.88$ & $32.57\pm1.19$ & $33.13\pm0.47$ \\
& Marmousi 2 & $21.24\pm1.88$ & $24.98\pm1.25$ & $22.01\pm1.56$ & $23.55\pm1.42$ & $22.89\pm1.63$ & $26.66\pm0.97$ & $27.12\pm0.53$ \\
\cline{1-9}
\multirow{2}{*}{\textbf{$R^2 \uparrow$}} & SEAM & $0.9715\pm0.0082$ & $0.9929\pm0.0053$ & $0.9747\pm0.0076$ & $0.9843\pm0.0066$ & $0.9836\pm0.0062$ & $0.9940\pm0.0035$ & $0.9953\pm0.0006$\\
& Marmousi 2 & $0.9428\pm0.0107$ & $0.9733\pm0.0064$ & $0.9506\pm0.0082$ & $0.9641\pm0.0089$ & $0.9584\pm0.0097$ & $0.9807\pm0.0043$ & $0.9841\pm0.0019$\\
\cline{1-9}
\multirow{2}{*}{\textbf{SSIM $\uparrow $}} & SEAM & $0.8222\pm0.0288$ & $0.9237\pm0.0182$ & $0.8664\pm0.0257$ & $0.9118\pm0.0235$ & $0.9171\pm0.0227$ & $0.9476\pm0.0153$ & $0.9524\pm0.0079$\\
& Marmousi 2 & $0.8023\pm0.0267$ & $0.9107\pm0.0153$ & $0.8351\pm0.0182$ & $0.8785\pm0.0207$ & $0.8689\pm0.0223$ & $0.9344\pm0.0127$ & $0.9388\pm0.0056$\\
\cline{1-9}
\multirow{2}{*}{\textbf{MAE $\downarrow$}} & SEAM & $0.1147\pm0.0181$ & $0.0533\pm0.0096$ & $0.0888\pm0.0165$ & $0.0552\pm0.0153$ & $0.0625\pm0.0142$ & $0.0451\pm0.0087$ & $0.0424\pm0.0046$\\
& Marmousi 2  & $0.1661\pm0.0228$ & $0.0995\pm0.0127$ & $0.1458\pm0.0153$ & $0.1083\pm0.0165$ & $0.1196\pm0.0182$ & $0.0792\pm0.0086$ & $0.0745\pm0.0041$\\
\cline{1-9}
\multirow{2}{*}{\textbf{MSE $\downarrow$}} & SEAM & $0.0263\pm0.0083$ & $0.0066\pm0.0028$ & $0.0241\pm0.0075$ & $0.0095\pm0.0056$ & $0.0118\pm0.0052$ & $0.0052\pm0.0027$ & $0.0046\pm0.0007$\\
& Marmousi 2 & $0.0576\pm0.0108$ & $0.0252\pm0.0062$ & $0.0474\pm0.0078$ & $0.0334\pm0.0087$ & $0.0407\pm0.0092$ & $0.0171\pm0.0047$ & $0.0155\pm0.0019$\\
\bottomrule[1.2pt]
\end{tabular}
\caption{The quantitative metrics of various methods on SEAM and Marmousi 2 datasets.}
\label{tab: comparison}
\end{table*}
\begin{table*}[!t]
\centering
\scriptsize
\setlength{\tabcolsep}{3.2mm}
\renewcommand\arraystretch{1}
\begin{tabular}{cc|ccc|ccc|ccc}
\toprule[1.2pt]
\multirow{2}{*}{\textbf{Metric}} & \multirow{2}{*}{\textbf{Dataset}} & \multicolumn{3}{c}{\textbf{SNR of seismic data}} & \multicolumn{3}{c}{\textbf{Seismic wavelet}} & \multicolumn{3}{c}{\textbf{Num of wells (labels)}} \\
\cline{3-11}
 &  & \textbf{5} & \textbf{10} & \textbf{15} & \textbf{Ricker} & \makecell[c]{\textbf{Generalized} \\ \cite{48}} & \makecell[c]{\textbf{Berlage} \\ \cite{49}} & \textbf{4} & \textbf{6} &\textbf{8}\\
\cline{1-11}
\multirow{2}{*}{\textbf{SNR $\uparrow $}} & SEAM & 27.2946 & 30.0369 & 30.8089 & 33.1335  & 36.1062 & 35.5610 & 22.5420 & 26.8512 & 32.1195\\
& Marmousi 2 & 21.8899 & 24.9624 & 25.7309 & 27.1238 & 27.2290 & 30.3897 & 21.6468 & 22.5282 & 25.3285\\
\cline{1-11}
\multirow{2}{*}{\textbf{$R^2 \uparrow$}} & SEAM & 0.9816 & 0.9902 & 0.9918 & 0.9953 & 0.9976 & 0.9973 & 0.9450 & 0.9796 & 0.9939\\
& Marmousi 2 & 0.9472 & 0.9740 & 0.9782 & 0.9841 & 0.9846 & 0.9925 & 0.9442 & 0.9544 & 0.9761\\
\cline{1-11}
\multirow{2}{*}{\textbf{SSIM $\uparrow $}} & SEAM & 0.7486 & 0.8642 & 0.8990 & 0.9524 & 0.9644 & 0.9608 & 0.8809 & 0.9169 & 0.9638\\
& Marmousi 2 & 0.6886 & 0.8251 & 0.8927 & 0.9388 & 0.9340 & 0.9692 & 0.9126 & 0.9117 & 0.9344\\
\cline{1-11}
\multirow{2}{*}{\textbf{MAE $\downarrow$}} & SEAM & 0.0858 & 0.0635 & 0.0569 & 0.0424 & 0.0307 & 0.0336 & 0.1296 & 0.0784 & 0.0503\\
& Marmousi 2  & 0.1509 & 0.1020 & 0.0895 & 0.0745 & 0.0768 & 0.0509 & 0.1631 & 0.1370 & 0.0929\\
\cline{1-11}
\multirow{2}{*}{\textbf{MSE $\downarrow$}} & SEAM & 0.0184 & 0.0097 & 0.0081 & 0.0046 & 0.0024 & 0.0026 & 0.0545 & 0.0203 & 0.0061\\
& Marmousi 2 & 0.0525 & 0.0258 & 0.0201 & 0.0155 & 0.0154 & 0.0072 & 0.0556 & 0.0460 & 0.0237\\
\bottomrule[1.2pt]
\end{tabular}
\caption{Robustness experiment results of the RD-SCL under different seismic data noise levels, different wavelet types, and varying numbers of well logs.}
\label{tab: robust}
\end{table*}

Table~\ref{tab: comparison} reports the mean and error bound of metrics for each method computed over five random seeds. On SEAM, RD-SCL achieves the highest mean SNR ($+$0.56dB over PICL) and SSIM, along with the lowest MAE and MSE. Its error bounds are consistently the smallest among all methods, indicating superior training stability. RD-SCL also outperforms the strongest supervised method EIF by 2.81dB in mean SNR while reducing MAE by over 20$\%$. On Marmousi 2, RD-SCL leads in all metrics, surpassing PICL by 0.46dB SNR with its error bound ($\pm$0.53) nearly half that of PICL ($\pm$0.97). 

For qualitative assessment, we take the results obtained with $random~seed=2026$ as an example. Figs.\ref{fig: SEAM} visualize the SEAM inversion profiles and absolute residuals with magnified views of complex zones (and the visualization of Marmousi 2 can be found in \textit{Supplementary Materials}). RD-SCL faithfully reconstructs fault zones and lateral lithological variations, confirming that the proposed regularized deconvolution and cross-learning mechanisms preserve physical consistency.

\subsection{Field Application}
\begin{table}[t]
\centering
\setlength{\tabcolsep}{0.02\columnwidth}
\renewcommand\arraystretch{0.8}
\begin{tabular}{cccccc}
\toprule[1.2pt]
Metric & EIF  & ADDIN & SSL & PICL & RD-SCL \\
\midrule
SNR $\uparrow $ & 29.71 & 29.77 & 28.24 & 33.10 & 34.07 \\
$R^2 \uparrow$ & 0.7853 & 0.7781 & 0.6752 & 0.8963 & 0.9203 \\
MAE $\downarrow$ & 0.3491 & 0.2943 & 0.4355 & 0.2412 & 0.1994 \\
MSE $\downarrow$ & 0.1845 & 0.1419 & 0.3194 & 0.0981 & 0.0618 \\
\bottomrule[1.2pt]
\end{tabular}
\caption{Validation results of various methods on blind well, the mean over five random seeds.}
\label{tab: field well 2}
\end{table}
To evaluate practical generalization, we further tested all methods on a field dataset. Three wells ($\approx$1$\%$ of total traces) are used for training, with the remaining well held out as a blind validation set. Taking $random~seed=2026$ as an example, Fig.~\ref{fig: field} displays the inversion profiles; PICL and RD-SCL yield the most coherent reconstructions, preserving lateral continuity and vertical resolution with thin layers clearly discernible. On the blind well (Table~\ref{tab: field well 2}), RD‑SCL outperforms all competing methods across every metric, achieving the highest SNR (34.07dB) and R$^2$ (0.9203), exceeding PICL by 0.97dB and 0.0251 respectively, and demonstrating superior fidelity in recovering fine impedance variations under realistic noise and structural complexity.

\subsection{Computational efficiency}

Fig.~\ref{fig: efficiency} compares per-run time and trainable parameters across methods. RD-SCL uses only 56.5k parameters, substantially fewer than EIF, SSL, and PICL but more than TCN and SSM. Its runtime on SEAM markedly faster than ADDIN ($-$60$\%$) and PICL ($-$24$\%$), and this efficiency is attributed to two factors: the proposed $\mathcal{D}$ provides a efficient solution without auxiliary networks, and the deliberately lightweight structure avoids the redundancy of network stacks.

\subsection{Robustness and Ablation Experiments}

To verify the stability of RD-SCL under extreme conditions, we conducted three robustness tests; Table~\ref{tab: robust} reports the mean results over multiple random seeds.

\textbf{Noise robustness.} We added Gaussian noise to the seismic records at three SNR levels (5, 10, and 15dB). RD‑SCL maintains consistently high accuracy: on SEAM, the SNR of the inverted impedance decreases from 33.13dB to 27.29dB, while R$^2$ remains above 0.98 throughout; a similar trend holds on Marmousi 2.

\textbf{Wavelet variability.} In addition to the standard Ricker wavelet, we tested the generalized wavelet and the Berlage wavelet. RD-SCL performs stably across all types, with R$^2$ ranging between 0.96 and 0.99 on both datasets, indicating that the dynamically estimated wavelet adapts to different source characteristics without prior tuning. Notably, this work is restricted to time-invariant wavelets; extending to the time-varying case remains an open challenge that would further exacerbate the ill-posedness of AII.

\textbf{Label sparsity.} Reducing the number of wells from 8 to 4 on SEAM causes SNR to drop from 32.12dB to 22.54dB, yet the result remains far above the fully supervised baseline. These results confirm that RD-SCL can handle extreme label scarcity.

\textbf{Ablation study.} Fig.~\ref{fig: ablation}(a) reports the contribution of each core component. Removing the cross-learning term causes a significant performance drop, confirming that the cross-consistency constraint is essential for exploiting unlabeled data. The symmetric design yields additional gains over a unidirectional variant, indicating that bidirectional consistency better regularizes training. We further examined the regularization order and penalty coefficient $\lambda$ (Fig.~\ref{fig: ablation}(b)). Since optimal settings vary across datasets, we selected first-order regularization with $\lambda=0.01$ as a balanced choice. 

\section{Conclusion}

We have presented RD-SCL, a semi-supervised framework for acoustic impedance imaging that simultaneously addresses latent wavelets and extreme label scarcity. Its two key innovations are regularized deconvolution operator and a symmetric cross-learning strategy. The operator provides efficient and stable closed-form latent variable estimation; building on this, the cross-learning enforces mutual consistency between labeled and unlabeled traces, effectively delivering unsupervised constraint for vast amounts of unlabeled data. Experiments demonstrate that RD-SCL outperforms state-of-the-art methods across all metrics, achieving substantial gains over the best competitor. Robustness tests confirm that the method maintains strong performance under severe noise, multiple wavelet types, and substantially reduced label counts. With only 56.5k learnable parameters and competitive runtime, RD-SCL offers a practical and physically consistent solution for promising field application. 

\section*{Acknowledgments}

This research is supported by the Chengdu University of Technology Postgraduate Innovative Cultivation Program. A long-term maintenance version of code will be uploaded to \url{https://github.com/lexiaoheng/Mariana}.

\bibliography{aaai2027}


\end{document}